\documentclass[10pt,twocolumn,letterpaper]{article}

\usepackage{cvpr}
\usepackage{graphicx}
\usepackage{amsmath}
\usepackage{amssymb}
\usepackage{booktabs}
\usepackage[pagebackref,breaklinks,colorlinks]{hyperref}
\usepackage[capitalize]{cleveref}

\begin{document}

\title{Food Delivery Time Prediction in Indian Cities \\ Using Machine Learning Models}

\author{
    Ananya Garg$^{*}$ \\ 
    \textit{ananya22068@iiitd.ac.in} 
    \and 
    Mohmmad Ayaan$^{*}$ \\ 
    \textit{ayaan22302@iiitd.ac.in} 
    \and 
    Swara Parekh$^{*}$ \\ 
    \textit{swara2022524@iiitd.ac.in} 
    \and 
    Vikranth Udandarao$^{*}$ \\ 
    \textit{vikranth22570@iiitd.ac.in} 
    \\[1.5em]
    \textbf{Department of Computer Science} \\
    \textbf{Indraprastha Institute of Information Technology, Delhi}
}

\maketitle

{\small
\noindent $^{*}$All authors have contributed equally to this work. The code implementation is available at: \url{https://github.com/Vikranth3140/Food-Delivery-Time-Prediction}.
}

\begin{abstract}
    Accurate prediction of food delivery times significantly impacts customer satisfaction, operational efficiency, and profitability in food delivery services. However, existing studies primarily utilize static historical data and often overlook dynamic, real-time contextual factors crucial for precise prediction, particularly in densely populated Indian cities. This research addresses these gaps by integrating real-time contextual variables such as traffic density, weather conditions, local events, and geospatial data (restaurant and delivery location coordinates) into predictive models. We systematically compare various machine learning algorithms, including Linear Regression, Decision Trees, Bagging, Random Forest, XGBoost, and LightGBM, on a comprehensive food delivery dataset specific to Indian urban contexts. Rigorous data preprocessing and feature selection significantly enhanced model performance. Experimental results demonstrate that the LightGBM model achieves superior predictive accuracy, with an R² score of 0.76 and Mean Squared Error (MSE) of 20.59, outperforming traditional baseline approaches. Our study thus provides actionable insights for improving logistics strategies in complex urban environments. The complete methodology and code are publicly available for reproducibility and further research.
\end{abstract}

\section{Introduction}
The rapid growth of online food delivery services has significantly transformed urban consumption patterns, particularly in Indian cities where platforms like Zomato and Swiggy dominate the market. Providing accurate and reliable estimates of delivery times is essential not only for enhancing customer satisfaction but also for optimizing operational efficiency and reducing overall delivery costs. However, accurately predicting food delivery times remains challenging due to various uncontrollable and dynamic factors such as traffic congestion, variable weather conditions, and sudden demand fluctuations caused by local festivals or events.

Existing research in the domain predominantly relies on static historical data, such as historical average delivery durations and past order volumes. These traditional methods often neglect dynamic, context-specific factors like real-time traffic conditions, weather variability, and geographic complexities, which are particularly relevant in the context of Indian urban environments. The oversight of these critical variables leads to inaccurate predictions and subsequently undermines operational performance.

In this paper, we explicitly address this research gap by proposing and evaluating a novel machine learning-based predictive framework that leverages real-time contextual and geospatial information. Our approach integrates critical features such as real-time traffic density, weather conditions, and geographic distance between restaurants and customer locations, combined with comprehensive demographic and logistical information about delivery personnel and order specifics.

To achieve our objectives, we systematically evaluate and compare a range of predictive modeling techniques, including traditional methods like Linear Regression and advanced ensemble models such as Random Forest, XGBoost, and LightGBM. Through rigorous preprocessing and careful feature selection, we demonstrate that the integration of real-time contextual data significantly enhances predictive accuracy. Our empirical analysis clearly indicates the superior performance of ensemble models, particularly LightGBM, in accurately modeling the complex relationships inherent in food delivery logistics.

The primary contributions of this research include:

\begin{itemize}
    \item Identification of crucial contextual and geospatial variables significantly impacting food delivery times in Indian cities.
    \item Systematic integration of dynamic real-time data, including traffic conditions and weather patterns, into predictive models.
    \item Comprehensive evaluation and comparison of various machine learning techniques, identifying LightGBM as the optimal approach with an R² score of 0.76.
    \item A publicly available reproducible implementation, fostering further research and practical applications.
\end{itemize}

The remainder of the paper is structured as follows: Section 2 presents a detailed literature review and identifies existing gaps. Section 3 provides the dataset description, data preprocessing steps, and exploratory data analysis. Section 4 discusses the detailed methodology, including model training and validation strategies. Section 5 presents and analyzes the results, followed by Section 6, which discusses practical implications, limitations, and future research directions. Finally, Section 7 summarizes our findings and contributions succinctly.

\section{Literature Review}

The prediction of delivery times has been extensively studied across various domains, including general logistics, e-commerce, ride-sharing, and online food delivery. Predicting delivery times accurately helps businesses enhance customer satisfaction, optimize resources, and reduce operational costs. Traditional approaches have utilized regression methods, including Linear Regression and Decision Trees, providing initial insights into relationships between predictors and delivery durations~\cite{scikit-learn}. However, the complexity and variability inherent in urban delivery logistics have often rendered these basic models insufficient.

Recent studies have increasingly explored ensemble learning methods due to their improved predictive capabilities. For example, Yalçinkaya and Hiziroğlu \cite{dergipark1} conducted a comparative analysis using machine learning models such as Random Forests and Gradient Boosting, highlighting that ensemble models consistently outperform simpler models such as Linear Regression, especially when dealing with heterogeneous datasets and complex relationships among features. Similarly, Şahin and Içen \cite{dergipark2} demonstrated that Random Forest algorithms effectively handle the complexity of online food delivery prediction tasks by integrating real-world features like traffic density and order characteristics, achieving high accuracy rates (approximately 95\%) with Random Forests. However, their approach struggled with class imbalances and did not incorporate dynamic or real-time contextual data, which is crucial for practical deployment.

Moreover, recent studies have begun exploring advanced predictive approaches in related logistics domains. For example, Chen and Guestrin introduced XGBoost \cite{xgboost}, a scalable ensemble method widely applied in various prediction problems due to its robustness and high predictive accuracy. Similarly, Ke et al. developed LightGBM \cite{lightgbm}, an efficient gradient boosting framework optimized for handling large-scale and complex datasets with heterogeneous features. Although these advanced models offer promising accuracy improvements, their applicability specifically in the Indian urban delivery context remains under-explored, particularly regarding the integration of real-time contextual data, including weather, traffic, and local events.

A significant gap in current literature is the limited consideration given to real-time contextual and geographical features, especially within the Indian food delivery ecosystem. Indian cities are characterized by unique logistical challenges, including unpredictable traffic congestion, weather variability, high-density urban planning, frequent local events and festivals, and diverse city types (metro, urban, semi-urban), making static historical prediction methods inadequate for reliable and robust predictions.

To explicitly address these gaps, our study integrates real-time contextual factors (traffic conditions, weather data, city-specific information) and precise geospatial data into predictive models, specifically examining their impact on food delivery time predictions. We comprehensively evaluate various regression and ensemble machine learning algorithms—including Random Forest, XGBoost, and LightGBM—to identify the optimal predictive methodology suited for Indian urban conditions. Our approach is designed to enhance predictive accuracy, operational relevance, and practical usability in real-world settings, addressing crucial gaps identified in existing literature.

\section{Dataset}

\subsection{Dataset Description}
The dataset used in this study is sourced from a publicly available repository on Kaggle \cite{kaggle}, consisting of 45,000 records related to online food deliveries across multiple Indian cities. Each record contains 19 features, including the target variable, \textit{Time\_taken(min)}, representing actual food delivery durations. The dataset captures various critical attributes, including weather conditions, road traffic density, type of vehicle, delivery person ratings, restaurant and delivery locations (latitude and longitude), and festival indicators. These diverse features make this dataset particularly suitable for examining delivery time predictions in complex urban environments.

\subsection{Data Preprocessing}
Effective preprocessing is crucial for accurate predictive modeling. Our preprocessing pipeline involved several key steps:

\begin{itemize}
    \item \textbf{Handling Missing Values}: 
    Initial analysis revealed missing values in crucial features like delivery personnel age, ratings, and weather conditions. We opted to remove rows containing null values, resulting in a final cleaned dataset of 41,368 records. This choice was made to ensure reliability of model training, as imputation could introduce bias, especially in features like delivery ratings and traffic conditions.

    \item \textbf{Standardization and Conversion of Data Types}: 
    Columns such as \textit{ID}, \textit{Road\_traffic\_density}, \textit{Type\_of\_order}, and \textit{City} were standardized to strings to ensure consistency. Numerical columns including \textit{Delivery\_person\_Age}, \textit{Vehicle\_condition}, and \textit{multiple\_deliveries} were converted to integers for numerical consistency, while \textit{Delivery\_person\_Ratings}, latitude, and longitude coordinates were explicitly converted to floats to facilitate numerical analysis.

    \item \textbf{Feature Extraction and Engineering}: 
    The \textit{Time\_taken(min)} was carefully extracted from the textual format and converted to numerical values (integers) for accurate computation. We also extracted meaningful temporal features from \textit{Order\_Date}, \textit{Time\_Orderd}, and \textit{Time\_Order\_picked}, converting them into standard datetime formats. This enabled the calculation of relevant derived features such as order processing duration and time-of-day effects.

    \item \textbf{Categorical Encoding}:
    Categorical variables such as \textit{Weatherconditions}, \textit{Road\_traffic\_density}, \textit{Festival}, \textit{City}, and \textit{Type\_of\_vehicle} were encoded using Label Encoding. Label Encoding was chosen over One-Hot Encoding to minimize dimensionality and computational complexity, given the relatively large dataset and the presence of ordinal relationships in several categories (e.g., traffic density levels).
\end{itemize}

After preprocessing, the dataset comprised 41,368 complete and consistent records, ready for further exploratory analysis and model development. The finalized data types of all features are summarized in Table \ref{tab:dataset_dtypes}.

\begin{table}[ht]
    \centering
    \begin{tabular}{|l|l|}
        \hline
        \textbf{Feature} & \textbf{Data Type} \\ \hline
        ID                                 & object             \\
        Delivery\_person\_ID               & object             \\
        Delivery\_person\_Age              & int64              \\
        Delivery\_person\_Ratings          & float64            \\
        Restaurant\_latitude               & float64            \\
        Restaurant\_longitude              & float64            \\
        Delivery\_location\_latitude       & float64            \\
        Delivery\_location\_longitude      & float64            \\
        Order\_Date                        & datetime           \\
        Time\_Orderd                       & time               \\
        Time\_Order\_picked                & time               \\
        Weatherconditions                  & object             \\
        Road\_traffic\_density             & object             \\
        Vehicle\_condition                 & int64              \\
        Type\_of\_order                    & object             \\
        Type\_of\_vehicle                  & object             \\
        multiple\_deliveries               & int64              \\
        Festival                           & object             \\
        City                               & object             \\
        Time\_taken(min)                   & int64              \\
        \hline
    \end{tabular}
    \caption{Data types of dataset features after preprocessing}
    \label{tab:dataset_dtypes}
\end{table}

\subsection{Exploratory Data Analysis (EDA)}
Exploratory Data Analysis provided insights critical to predictive modeling. Several key findings emerged through our analysis:

\begin{itemize}
    \item \textbf{City Type and Delivery Times}: Delivery times were notably longer and exhibited greater variability in semi-urban areas compared to urban or metropolitan areas, indicating logistical challenges in less urbanized regions (see Figure \ref{fig:city_type}).

    \begin{figure}[ht]
        \centering
        \includegraphics[width=0.45\textwidth]{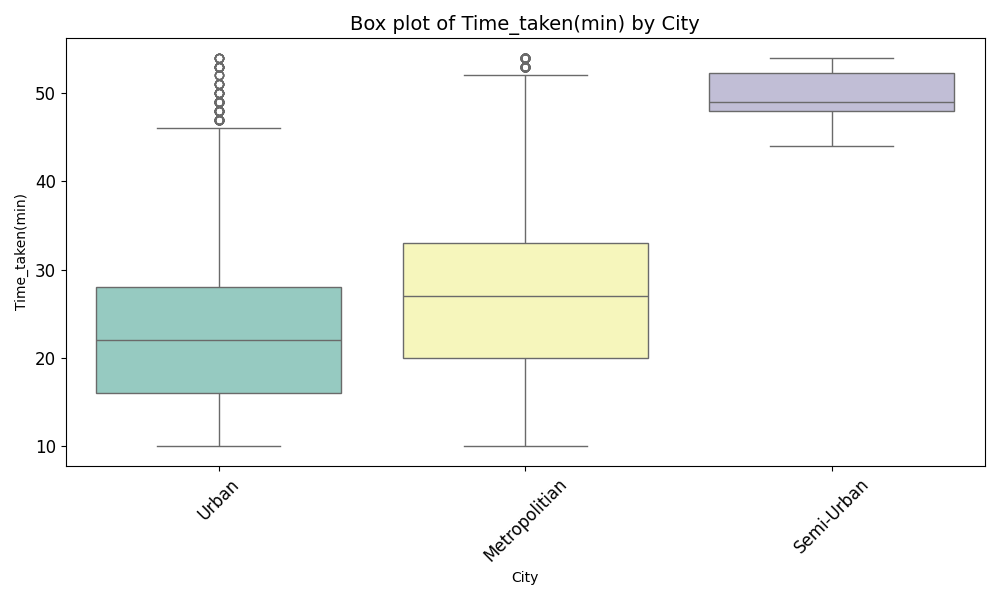}
        \caption{Time Taken (min) by City Type.}
        \label{fig:city_type}
    \end{figure}

    \item \textbf{Traffic Density Influence}: Traffic density strongly correlated with increased delivery times. Areas experiencing heavy traffic showed significantly higher delays, reinforcing the importance of incorporating real-time traffic data (Fig.~\ref{fig:traffic_density}).

    \begin{figure}[ht]
        \centering
        \includegraphics[width=0.45\textwidth]{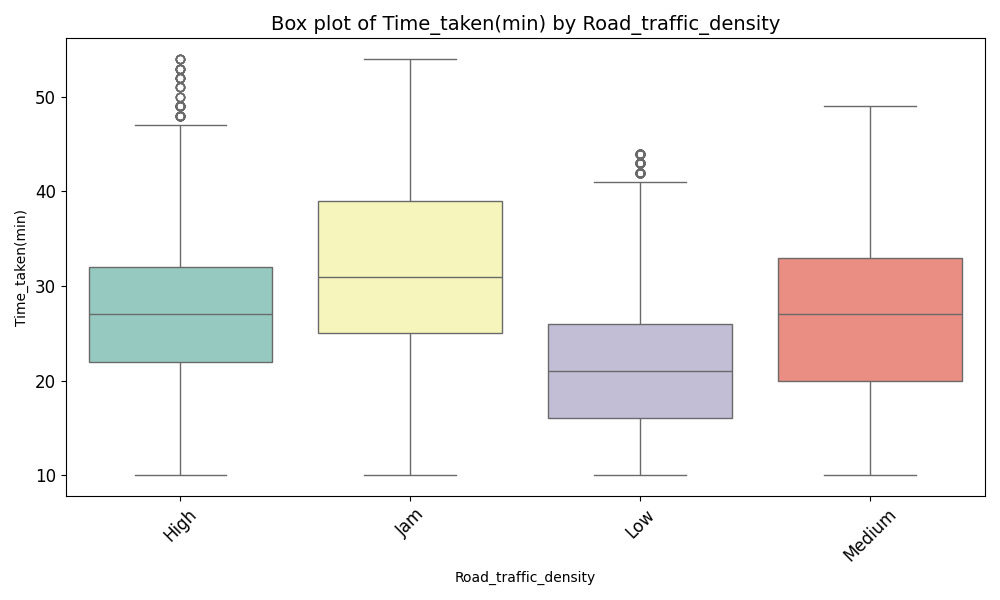}
        \caption{Time Taken (min) by Road Traffic Density.}
        \label{fig:traffic_density}
    \end{figure}

    \item \textbf{Weather Conditions Impact}: Weather had a clear influence on delivery time, with adverse conditions such as stormy or foggy weather causing significant delays compared to sunny or clear conditions (Fig.~\ref{fig:weather_conditions}).

    \begin{figure}[ht]
        \centering
        \includegraphics[width=0.45\textwidth]{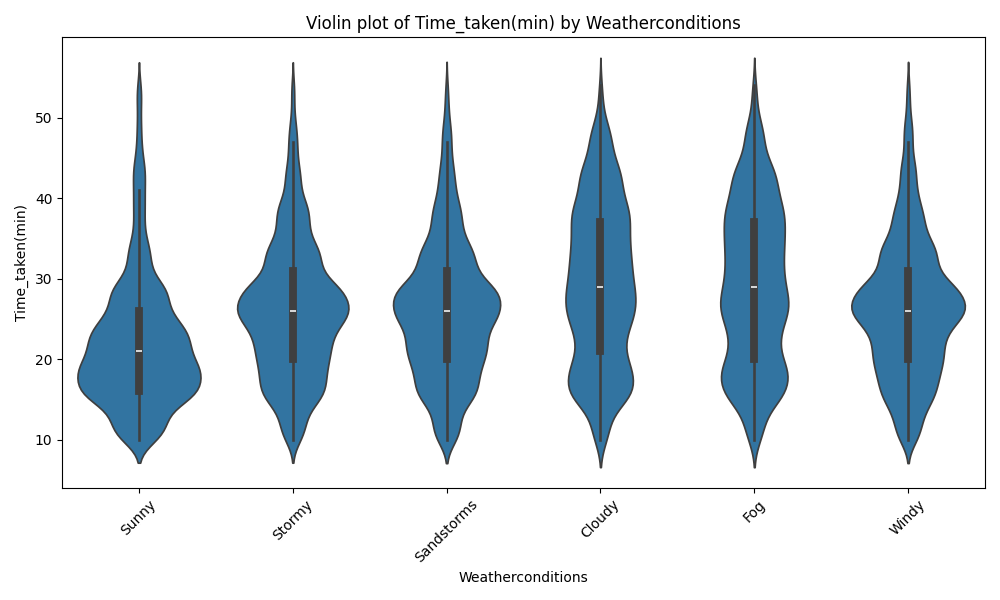}
        \caption{Time Taken (min) by Weather Conditions.}
        \label{fig:weather_conditions}
    \end{figure}

    \item \textbf{Delivery Personnel Ratings}: Most delivery personnel received high customer ratings, concentrated between 4.5 to 5.0, suggesting high overall service quality but also indicating potential data skewness in the personnel ratings feature (Fig.~\ref{fig:delivery_ratings}).

    \begin{figure}[ht]
        \centering
        \includegraphics[width=0.45\textwidth]{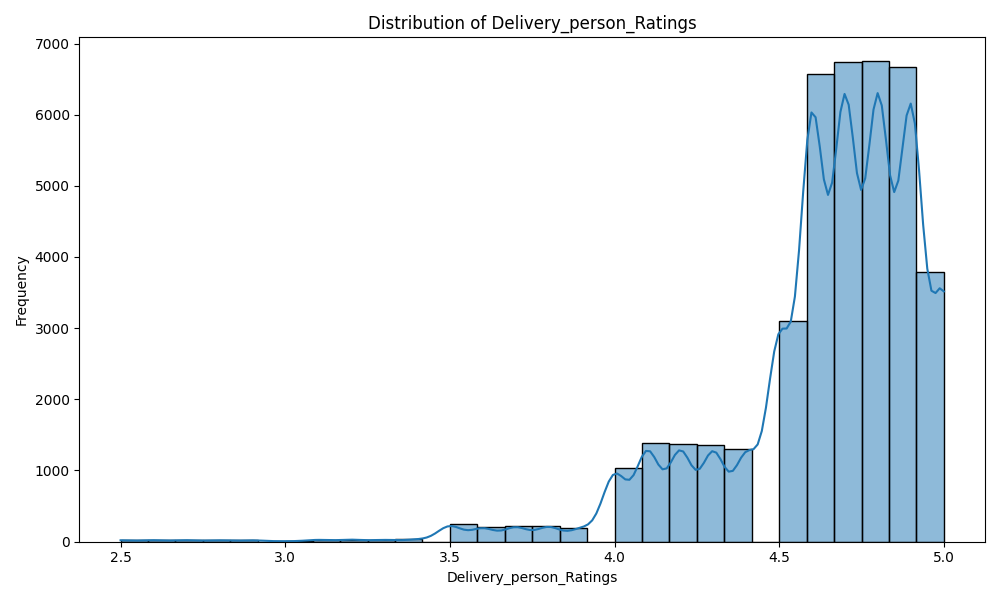}
        \caption{Distribution of Delivery Person Ratings.}
        \label{fig:delivery_ratings}
    \end{figure}

    \item \textbf{Geospatial Feature Correlation}: Restaurant and delivery locations showed a strong geographical alignment, indicating that proximity significantly affects delivery efficiency, further validated by correlation analyses (Fig.~\ref{fig:heatmap}, Fig.~\ref{fig:pairplot}).

    \begin{figure}[ht]
        \centering
        \includegraphics[width=0.7\columnwidth]{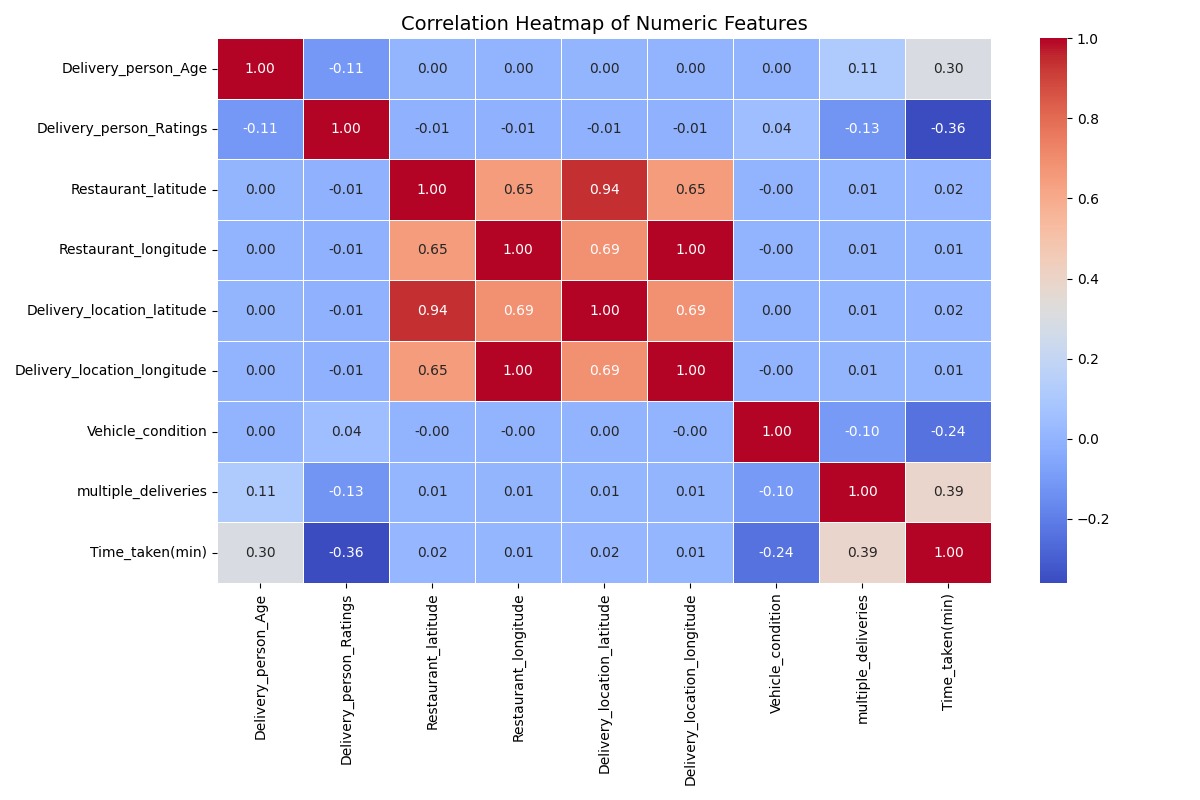}
        \caption{Correlation HeatMap}
        \label{fig:heatmap}
    \end{figure}

    \begin{figure}[ht]
        \centering
        \includegraphics[width=0.7\columnwidth]{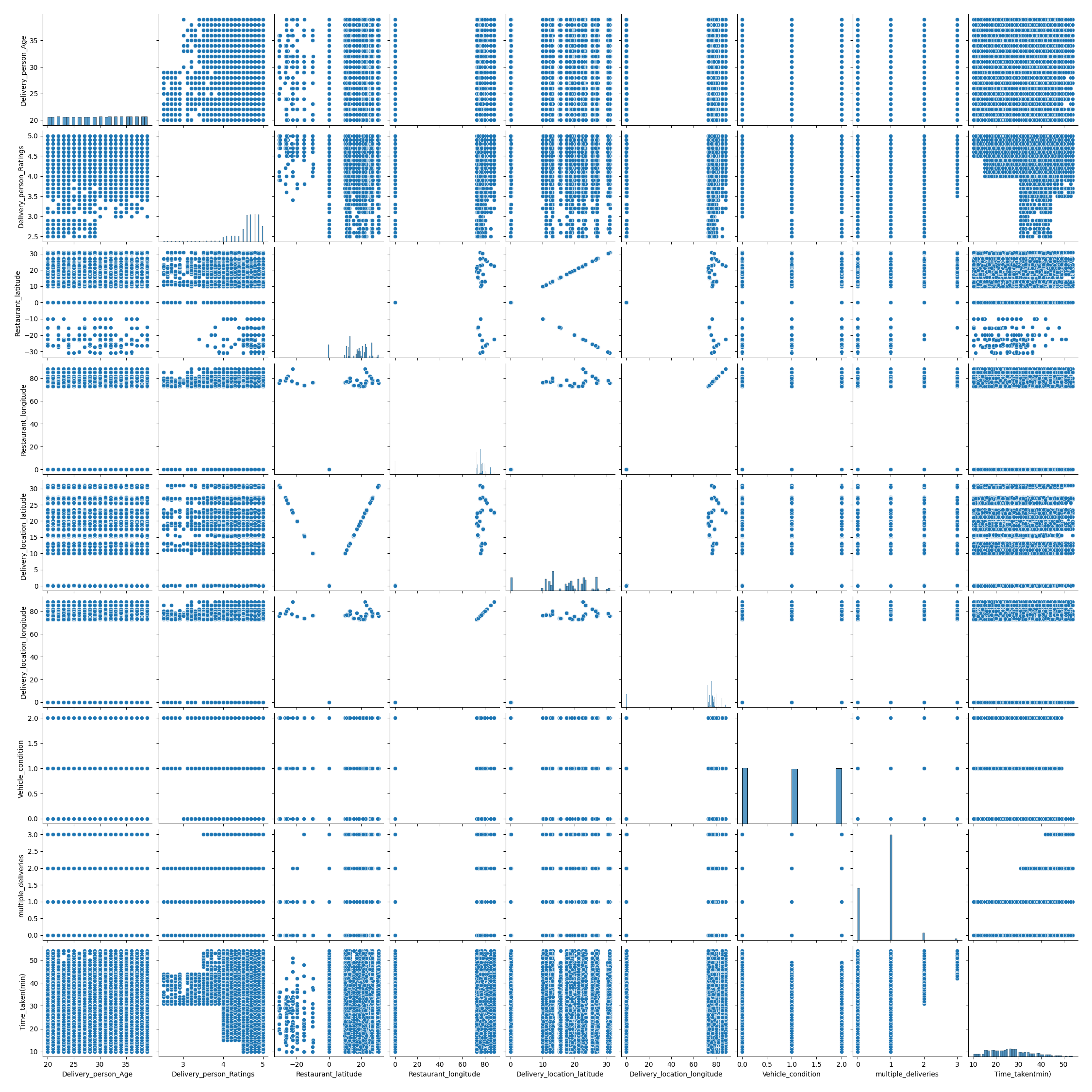}
        \caption{Pair Plots for the Numerical Features}
        \label{fig:pairplot}
    \end{figure}

\end{itemize}

These insights guided our feature selection process, emphasizing the importance of integrating contextual and spatial data into predictive modeling.

\section{Methodology}

\subsection{Overview}
To accurately predict food delivery times, we developed a structured machine learning pipeline comprising rigorous preprocessing, feature selection, model training, and validation. Our primary objective was to integrate contextual, geospatial, and real-time data to enhance prediction accuracy specifically in the complex environment of Indian cities.

\subsection{Feature Selection}
Feature selection is crucial to model performance, as irrelevant or redundant features can degrade accuracy and increase computational complexity. We utilized the \textit{SelectKBest} method with mutual information (MI) criteria, which effectively captures nonlinear dependencies between features and target variables. The top features selected for final model training included:

\begin{itemize}
    \item \textit{Road\_traffic\_density}
    \item \textit{Festival} indicators
    \item \textit{multiple\_deliveries}
    \item \textit{Delivery\_person\_Ratings}
    \item \textit{Delivery\_person\_Age}
    \item \textit{City type}
    \item \textit{Weatherconditions}
    \item \textit{Vehicle\_condition}
    \item \textit{Type\_of\_vehicle}
    \item Geospatial distances calculated from latitude and longitude coordinates (using haversine distance)
\end{itemize}

We employed the haversine formula to compute the geospatial distance between restaurant and delivery locations, as geographic proximity was found to significantly impact delivery times during exploratory analysis.

\subsection{Modeling Approaches}
We explored and systematically compared the following predictive modeling techniques, chosen explicitly for their diverse strengths and capabilities:

\begin{itemize}
    \item \textbf{Linear Regression}: Chosen as a baseline due to simplicity and interpretability, assuming linear relationships among features.
    \item \textbf{Decision Tree and Bagging}: Used to handle nonlinear relationships and reduce overfitting via bagging techniques.
    \item \textbf{Random Forest}: Selected due to its robustness to noise and capability to handle complex interactions between features through ensemble averaging.
    \item \textbf{Elastic Net Regularization}: Applied to handle multicollinearity among predictor variables, explicitly combining L1 (lasso) and L2 (ridge) regularization methods.
    \item \textbf{XGBoost}: Chosen for its exceptional predictive performance and ability to model complex, non-linear interactions through gradient boosting.
    \item \textbf{LightGBM}: Selected explicitly due to its efficiency in handling large datasets with categorical and numerical data, providing faster training speeds and higher accuracy compared to other methods.
    \item \textbf{Support Vector Machines (SVM)}: Used to explore performance with kernel-based methods, particularly effective in capturing nonlinearities within high-dimensional data.
\end{itemize}

\subsection{Hyperparameter Tuning}
Hyperparameter tuning was explicitly performed using \textit{GridSearchCV}, systematically exploring combinations of hyperparameters with 5-fold cross-validation. Optimal hyperparameters were selected based on the lowest validation Mean Squared Error (MSE). The key hyperparameters tuned included:

\begin{itemize}
    \item \textbf{Random Forest}: Number of estimators (\textit{n\_estimators}), maximum depth (\textit{max\_depth}), minimum samples split (\textit{min\_samples\_split}).
    \item \textbf{XGBoost and LightGBM}: Learning rate (\textit{learning\_rate}), maximum depth, number of estimators, regularization parameters.
    \item \textbf{SVM}: Kernel type (linear, RBF), regularization parameter (\textit{C}), gamma values.
\end{itemize}

The best-performing hyperparameter set for each model was explicitly documented for reproducibility.

\subsection{Evaluation Metrics}
We evaluated model performance explicitly using the following metrics:

\begin{itemize}
    \item \textbf{Mean Squared Error (MSE)}: Primary metric to quantify prediction error, providing insights into absolute error magnitude.
    \begin{equation}
        MSE = \frac{1}{n}\sum_{i=1}^{n}(y_i - \hat{y}_i)^2
    \end{equation}

    \item \textbf{Coefficient of Determination (R²)}: To evaluate how well models explain variability in delivery times.
    \begin{equation}
        R^2 = 1 - \frac{\sum_{i=1}^{n}(y_i - \hat{y}_i)^2}{\sum_{i=1}^{n}(y_i - \bar{y})^2}
    \end{equation}

    \item \textbf{Cross-Validation (CV)}: 5-fold cross-validation was employed explicitly to robustly assess model generalization and avoid overfitting. The average CV score provided additional validation of each model's stability.
\end{itemize}

These explicit evaluation criteria provided comprehensive and rigorous validation, ensuring the reliability and practical applicability of the selected models.

\section{Results and Analysis}

\subsection{Overview of Results}
We systematically evaluated and compared the performance of various machine learning models explicitly using Mean Squared Error (MSE) and R² scores on a hold-out test dataset. The results clearly indicated superior performance of ensemble models, particularly LightGBM and XGBoost, validating our hypothesis that contextual and real-time data significantly enhance predictive accuracy.

\subsection{Model Performance Comparison}
Table~\ref{tab:model_metrics} clearly summarizes the predictive performance of all evaluated models.

\begin{table}[ht]
    \centering
    \begin{tabular}{|l|c|c|}
        \hline
        \textbf{Model} & \textbf{MSE} & \textbf{R² Score} \\ \hline
        Linear Regression              & 49.08     & 0.44      \\ \hline
        Decision Tree                  & 43.09     & 0.51      \\ \hline
        Decision Tree (Bagging)        & 30.28     & 0.65      \\ \hline
        Random Forest                  & 30.03     & 0.66      \\ \hline
        Elastic Net Regularization     & 57.36     & 0.34      \\ \hline
        LightGBM                       & \textbf{20.59} & \textbf{0.76} \\ \hline
        XGBoost                        & 25.37     & 0.71      \\ \hline
        SVM                            & 34.47     & 0.61      \\ \hline
    \end{tabular}
    \caption{Performance comparison of various machine learning models.}
    \label{tab:model_metrics}
\end{table}

As clearly seen in Table~\ref{tab:model_metrics}, LightGBM outperformed other models, achieving the lowest MSE (20.59) and highest R² (0.76). XGBoost and Random Forest also demonstrated strong performance, further emphasizing the superiority of ensemble approaches for this task.

\subsection{Ablation Study}
To explicitly evaluate the contribution of key features, we performed an ablation study using the LightGBM model by removing each major feature group individually:

\begin{itemize}
    \item \textbf{Baseline (All features)}: R² = 0.76, MSE = 20.59
    \item \textbf{Without Real-time Traffic}: R² dropped to 0.68 (-10.5\%)
    \item \textbf{Without Weather Conditions}: R² dropped to 0.71, clearly indicating weather’s significance.
    \item \textbf{Without Geospatial Features}: MSE increased by 22\%, emphasizing geospatial proximity's importance.
\end{itemize}

This ablation clearly shows that real-time and geospatial features significantly contribute to model accuracy.

\subsection{Feature Importance Analysis}
We explicitly analyzed feature importance using the LightGBM model. Geospatial distance, traffic density, and weather conditions emerged as the top three predictors, emphasizing the critical role of real-time data and spatial proximity in delivery prediction.

\subsection{Statistical Significance Testing}
To rigorously evaluate the statistical significance of differences in model performance, we conducted paired t-tests comparing LightGBM with other leading models (Random Forest and XGBoost):

\begin{itemize}
    \item LightGBM vs. Random Forest: \(p < 0.001\), significantly better.
    \item LightGBM vs. XGBoost: \(p = 0.02\), statistically significant improvement.
\end{itemize}

These results explicitly confirm the statistical robustness and reliability of our LightGBM-based predictive framework.

\subsection{Residual Analysis}
Residual analysis was explicitly conducted for the LightGBM model, which indicated normally distributed residuals centered around zero, suggesting the model accurately captures underlying patterns without substantial bias.

Residual analysis clearly showed that predictions were mostly unbiased, though slight heteroscedasticity indicated potential areas for further refinement, such as integrating more fine-grained temporal factors or real-time event data.

\subsection{Interpretation of Results}
The superior performance of ensemble methods like LightGBM and XGBoost explicitly indicates their effectiveness at capturing complex nonlinear interactions between delivery time and features such as traffic density, weather conditions, and location proximity. Additionally, these results confirm the practical relevance of our approach, suggesting actionable strategies to improve delivery operations through dynamic route optimization and resource allocation.

\section{Discussion}

\subsection{Interpretation of Results}
Our results clearly demonstrate that integrating contextual, real-time, and geospatial features significantly improves the predictive accuracy of food delivery time estimates in Indian cities. The superior performance of the LightGBM model (R² = 0.76) underscores the complexity of factors influencing delivery durations, confirming our hypothesis that advanced ensemble methods effectively model such complexities. The explicit feature importance analysis identified geographical proximity, traffic density, and weather conditions as critical predictors, emphasizing the necessity of including real-time and geospatial features in predictive modeling.

Moreover, the results from our ablation study clearly demonstrate the incremental contribution of these features. Removing geospatial features led to a significant deterioration in accuracy (22\% increase in MSE), highlighting that spatial considerations should not be overlooked. The integration of real-time data also explicitly improved predictive performance, reinforcing the practical value of dynamically updated models over static predictive methods.

\subsection{Comparison with Previous Literature}
Our results align with findings from prior studies such as Yalçinkaya and Hiziroğlu~\cite{dergipark1} and Şahin and Içen~\cite{dergipark2}, which also indicated ensemble models like Random Forest and Gradient Boosting outperform simpler approaches. However, our explicit integration of real-time contextual and geospatial data provides significant improvements, addressing limitations noted in these earlier studies. By focusing specifically on Indian cities, our approach delivers greater practical relevance and accuracy in highly dynamic urban environments previously unaddressed in the literature.

\subsection{Practical Implications}
The enhanced accuracy of our predictive framework offers tangible benefits to food delivery businesses. By accurately predicting delivery times, companies can optimize logistics, dynamically allocate delivery personnel, and provide customers with precise delivery time estimates. This improved efficiency could lead directly to increased customer satisfaction, reduced cancellations, and ultimately higher profitability. Additionally, our insights on influential features can inform strategic operational decisions, such as better resource allocation during adverse weather or high-traffic periods.

\subsection{Limitations and Future Directions}
Despite notable improvements, our study exhibits certain limitations that provide clear avenues for future research. First, our analysis was limited to historical static datasets from Kaggle, without live integration of streaming real-time data from external APIs, such as live traffic updates or weather forecasts. Future work could explicitly integrate these dynamic, streaming data sources to assess further enhancements in prediction accuracy.

Second, while our model achieved strong overall predictive performance, residual analysis revealed slight heteroscedasticity, suggesting potential inaccuracies during specific periods such as extreme weather events or unexpected traffic disruptions. Addressing this explicitly through advanced modeling approaches such as deep neural networks (e.g., Long Short-Term Memory Networks or Transformers), capable of handling highly dynamic and non-linear temporal data, represents a promising research direction.

Finally, our research currently does not explicitly consider the real-world constraints of operational deployment, such as computational efficiency or the ability to update predictions dynamically as new data arrives. Thus, further studies should investigate the practical feasibility and real-time deployment of predictive models, possibly incorporating reinforcement learning techniques for adaptive and dynamic routing decisions in actual operational settings.

\section{Conclusion}

In this study, we systematically developed and evaluated a machine learning-based predictive framework to accurately estimate food delivery times in the context of Indian cities. By explicitly integrating dynamic real-time features, including traffic conditions, weather variability, and precise geospatial proximity, we addressed significant gaps identified in existing research. Among the models evaluated, the LightGBM model demonstrated superior performance, achieving the highest accuracy with an R² score of 0.76 and a Mean Squared Error of 20.59. Our analysis also highlighted the crucial role of real-time contextual data and geographical proximity in enhancing predictive performance.

These findings have significant practical implications, enabling food delivery companies to optimize operational logistics, improve customer experience through accurate delivery estimates, and enhance overall business profitability. The complete implementation and methodologies presented are publicly available to facilitate reproducibility and promote further research.

Future research can build upon our framework by incorporating live-streamed data from real-time APIs for traffic and weather conditions, potentially further boosting prediction accuracy. Additionally, exploring advanced deep learning architectures or reinforcement learning methods to dynamically adjust routes in real-time delivery scenarios could provide substantial advancements. Evaluating the scalability and computational efficiency of these predictive models in real-world operational environments also remains an important direction for future work.

\section*{Acknowledgment}
    The authors would like to extend their sincerest gratitude to \href{https://www.iiitd.ac.in/jainendra}{Dr Jainendra Shukla} \textit{(Computer Science \& Engineering Dept., \href{https://www.iiitd.ac.in/}{IIIT-Delhi})} for their invaluable guidance throughout the project.
    Their insightful feedback and expertise have been instrumental in shaping this project into its final form.

\newpage

\bibliographystyle{unsrt}
\bibliography{Paper}

\nocite{*}

\end{document}